\pdfoutput=1
\documentclass[11pt,a4paper]{article}
\usepackage[hyperref]{acl2021}
\usepackage{times}
\usepackage{latexsym}

\usepackage{microtype}

\aclfinalcopy % Uncomment this line for the final submission
\usepackage{graphicx}
\usepackage{caption} 
\usepackage{subcaption}
\usepackage{stfloats}

\usepackage{hyperref}

\usepackage{appendix}

\usepackage{listings}
\usepackage{booktabs}  
\usepackage{multirow}

\usepackage[ruled,vlined]{algorithm2e}

\title{Data Augmentation for Sign Language Gloss Translation}
\author{\underline{Amit Moryossef}$^1$, \underline{Kayo Yin}$^3$,  Graham Neubig$^3$, Yoav Goldberg$^{1,2}$ \\
\texttt{amitmoryossef@gmail.com, kayoy@cs.cmu.edu}\\ 
\texttt{gneubig@cs.cmu.edu, yogo@cs.biu.ac.il}  \\
$^1$Bar-Ilan University   $^2$Allen Institute for AI\\
$^3$Language Technologies Institute, Carnegie Mellon University}

\begin{document}
\maketitle
\begin{abstract}

Sign language translation (SLT) is often decomposed into \textit{video-to-gloss} recognition and \textit{gloss-to-text} translation, where a gloss is a sequence of transcribed spoken-language words in the order in which they are signed. We focus here on gloss-to-text translation, which we treat as a low-resource neural machine translation (NMT) problem. However, unlike traditional low-resource NMT, gloss-to-text translation differs because gloss-text pairs often have a higher lexical overlap and lower syntactic overlap than pairs of spoken languages.
We exploit this lexical overlap and handle syntactic divergence by proposing two rule-based heuristics that generate pseudo-parallel gloss-text pairs from monolingual spoken language text. By pre-training on the thus obtained synthetic data, we improve translation from American Sign Language (ASL) to English and German Sign Language (DGS) to German by up to 3.14 and 2.20 BLEU, respectively.

\end{abstract}

\section{Introduction}\label{sec:intro}

Sign language is the most natural mode of communication for the Deaf.
However, in a predominantly hearing society, they often resort to lip-reading, text-based communication, or closed-captioning to interact with others.
Sign language translation (SLT) is an important research area that aims to improve communication between signers and non-signers while allowing each party to use their preferred language. SLT consists of translating a sign language (SL) video into a spoken language (SpL) text, and current approaches often decompose this task into two steps: (1) \emph{video-to-gloss}, or continuous sign language recognition (CSLR) \cite{cui2017recurrent, Camgoz_2018_CVPR}; (2) \emph{gloss-to-text}, which is a text-to-text machine translation (MT) task \cite{Camgoz_2018_CVPR,yin2020better}.

In this paper, we focus on gloss-to-text translation.
SL data and resources are often scarce, or nonexistent (\S\ref{sec:background}; \citet{bragg2019sign}). Gloss-to-text translation is, therefore, an example of an extremely low-resource MT task. However, while there is extensive literature on low-resource MT between spoken languages \cite{sennrich-etal-2016-improving,zoph2016transfer,xia2019generalized,zhou-etal-2019-handling}, the dissimilarity between sign and spoken languages calls for novel methods.
Specifically, as SL glosses borrow the lexical elements from their ambient spoken language, handling syntax and morphology poses greater challenges than lexeme translation (\S\ref{sec:whysl}).

\begin{figure}[t]
    \begin{center}
    \begin{subfigure}[t]{\linewidth}
        \includegraphics[width=\linewidth]{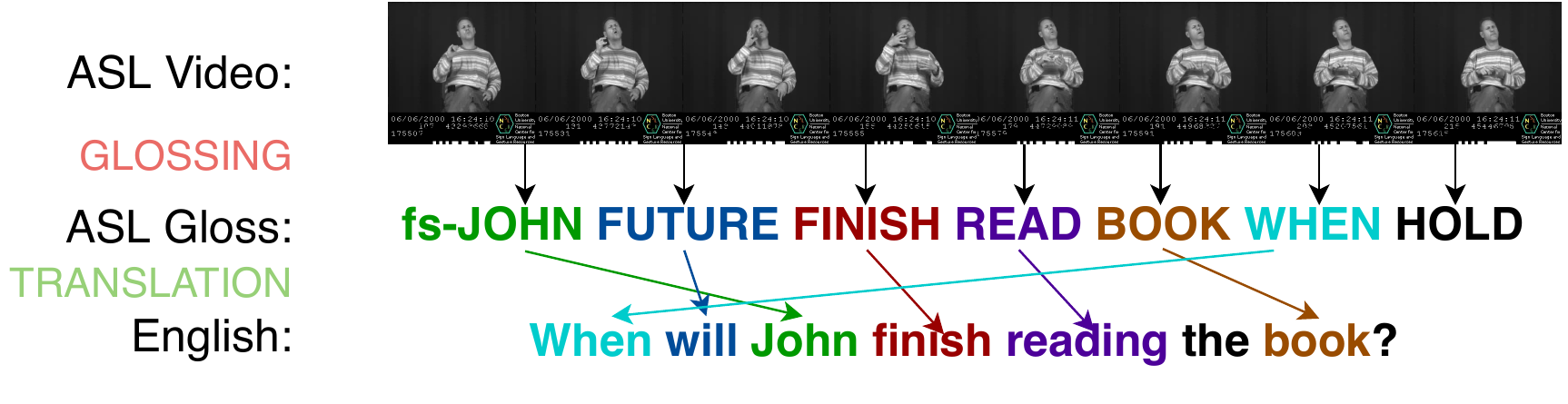}
        \caption{ASL video with gloss annotation and English translation}
        \label{fig:real-data}
    \end{subfigure}
    \par\bigskip
    \begin{subfigure}[t]{\linewidth}
        \includegraphics[width=\linewidth]{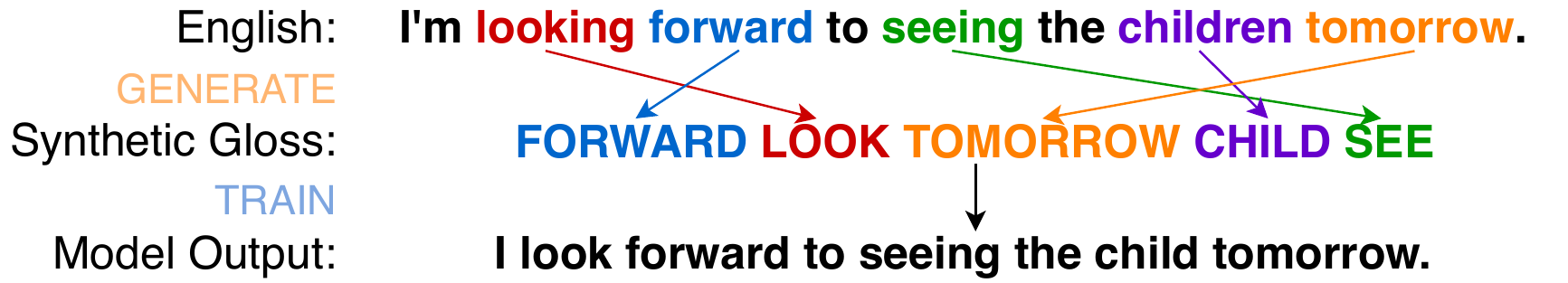}
        \caption{Data augmentation and training}
        \label{fig:fake-data}
    \end{subfigure}
    \end{center}
    \caption{Real and synthetic gloss-spoken pairs.}
    \label{fig:res}
\end{figure}

In this work, we 
(1) discuss the scarcity of SL data and quantify how the relationship between a sign and spoken language pair is different from a pair of two spoken languages; 
(2) show that the \textit{de facto} method for data augmentation using back-translation is not viable in extremely low-resource SLT; 
(3) propose two rule-based heuristics that exploit the lexical overlap and handles the syntactic divergence between sign and spoken language, to synthesize pseudo-parallel gloss/text examples (Figure \ref{fig:fake-data}); 
(4) demonstrate the effectiveness of our methods on two sign-to-spoken language pairs.

% Main figure

\section{Background}\label{sec:background} 

%\subsection{Sign Language Translation}

\paragraph{Sign Language Glossing}
SLs are often transcribed word-for-word using a spoken language through \emph{glossing} to aid in language learning, or automatic sign language processing \cite{ormel2010glossing}. While many SL glosses are words from the ambient spoken language, glossing preserves SL's original syntactic structure and therefore differs from translation (Figure \ref{fig:real-data}).

\paragraph{Data Scarcity}
While standard machine translation architectures such as the Transformer \cite{transformer} achieve reasonable performance on gloss-to-text datasets \cite{yin2020attention,camgoz2020sign}, parallel SL and spoken language corpora, especially those with gloss annotations, are usually far more scarce when compared with parallel corpora that exist between many spoken languages (Table \ref{table:g2tdata}). 

\begin{table*}[htb]
\resizebox{\textwidth}{!}{\begin{tabular}{lccc}
\toprule
& Language Pair & \# Parallel Gloss-Text Pairs & Vocabulary Size (Gloss / Spoken) \\
\midrule
Signum \cite{Agris2007TowardsAV} & DGS-German & 780 & 565 / 1,051 \\
NCSLGR \cite{ncslgr} & ASL-English & 1,875 & 2,484 / 3,104  \\
% British Sign Language Corpus \cite{schembri2013building} & British SL-English &\fix{@@} &\fix{@@} \\
% Swedish Sign Language Corpus \cite{mesch2015gloss} & Swedish SL-Swedish& & \\
RWTH-PHOENIX-Weather 2014T \cite{Camgoz_2018_CVPR} & DGS-German & 7,096 + 519 + 642 & 1,066 / 2,887 + 393 / 951 + 411 / 1,001 \\
Dicta-Sign-LSF-v2 \cite{11403/dicta-sign-lsf-v2/v1} & French SL-French & 2,904 & 2,266 / 5,028 \\
The Public DGS Corpus \cite{dataset:hanke-etal-2020-extending} & DGS-German & 63,912 & 4,694 / 23,404 \\
% KETI \cite{keti} & Korean SL-Korean & 105 & - \\
%Auslan Corpus & Auslan-English & 382 annotated videos? & \\
% Corpus of Finnish Sign Language \cite{salonen-etal-2020-corpus} & Finnish SL-Finnish &  & \\
\bottomrule
\end{tabular}}
\caption{Some publicly available SL corpora with gloss annotations and spoken language translations.} 
\label{table:g2tdata}
\end{table*}

\section{Sign vs. Spoken Language} \label{sec:whysl}
% \yg{this can be moved outside of the background, IMO, into its own section. Also, maybe rename it to "Sign language vs. Spoken Language" or something like that.}

Due to the paucity of parallel data for gloss-to-text translation, we can treat it as a low-resource translation problem and apply existing techniques for improving accuracy in such settings.
However, we argue that the relationship between glossed SLs and their spoken counterparts is different from the usual relationship between two spoken languages.
Specifically, glossed SLs are \emph{lexically similar but syntactically different} from their spoken counterparts. This contrasts heavily with the relationship among spoken language pairs where lexically similar languages tend also to be syntactically similar the great majority of the time.

To demonstrate this empirically, we adopt measures from \newcite{lin-etal-2019-choosing} to measure the lexical and syntactic similarity between languages, two features also shown to be positively correlated with the effectiveness of performing cross-lingual transfer in MT.

\paragraph{Lexical similarity} between two languages is measured using word overlap: 
$$
o_w = \frac{|T_1 \cap T_2|}{|T_1| + |T_2|}
$$
where $T_1$ and $T_2$ are the sets of types in a corpus for each language.
The word overlap between spoken language pairs is calculated using the TED talks dataset \cite{qi-etal-2018-pre}. The overlap between sign-spoken language pairs is calculated from the corresponding corpora in Table \ref{table:g2tdata}.

\begin{figure}[t]
    \begin{center}
    \includegraphics[width=\linewidth]{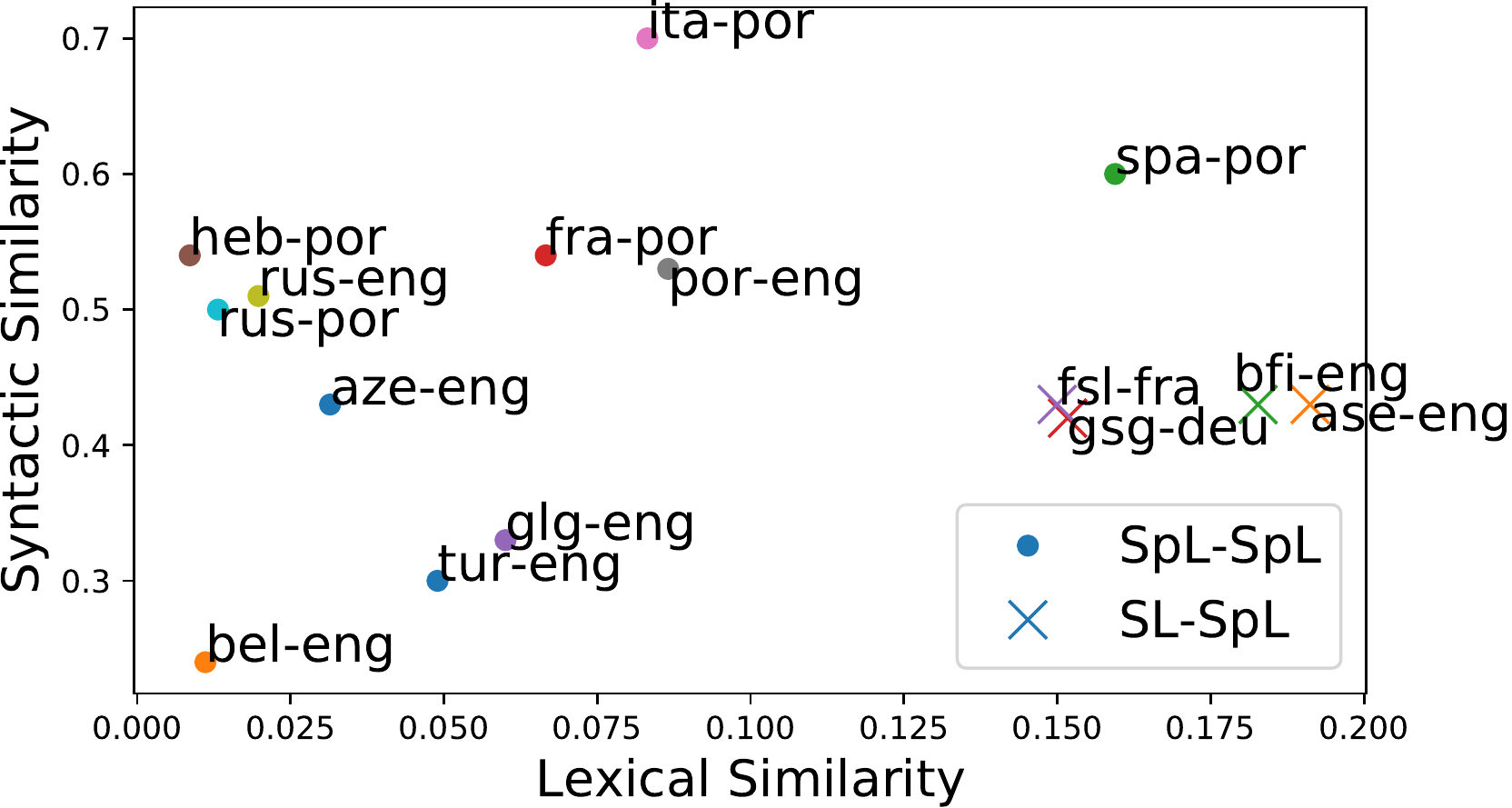}
    \end{center}
    \caption{Lexical and syntactic similarity between different language pairs denoted by their ISO639-2 codes.}
    \label{fig:overlaps}
\end{figure}

\paragraph{Syntactic similarity} between two languages is measured by $1-d_{syn}$ where $d_{syn}$ is the syntactic distance from \newcite{littell-etal-2017-uriel} calculated by taking the cosine distance between syntactic features adapted from the World Atlas of Language Structures \cite{wals}.

Figure \ref{fig:overlaps} shows that sign-spoken language pairs are indeed outliers with lower syntactic similarity and higher lexical similarity. 
We seek to leverage this fact and the high availability of monolingual spoken language data to compensate for the scarcity of SL resources. In the following section, we propose data augmentation techniques using word order modifications to create synthetic sign gloss data from spoken language corpora.

\section{Data Augmentation}\label{sec:methods}

This section discusses methods to improve gloss-to-text translation through data augmentation, specifically those that take monolingual corpora of standard spoken languages and generate pseudo-parallel ``gloss'' text.
We first discuss a standard way of doing so, back-translation, point out its potential failings in the SL setting, then propose a novel rule-based data augmentation algorithm.

\subsection{Back-translation}
Back-translation \cite{irvine-callison-burch-2013-combining, sennrich-etal-2016-improving} automatically creates pseudo-parallel sentence pairs from monolingual text to improve MT in low-resource settings.
However, back-translation is only effective with sufficient parallel data to train a functional MT model, which is not always the case in extremely low-resource settings \citep{currey-etal-2017-copied}, and particularly when the domain of the parallel training data and monolingual data to be translated are mismatched \citep{dou-etal-2020-dynamic}.

\subsection{Proposed Rule-based Augmentation Strategies}

Given the limitations of standard back-translation techniques, we next move to the proposed method of using rule-based heuristics to generate SL glosses from spoken language text.

\paragraph{General rules}
The differences in SL glosses from spoken language can be summarized by (1) A lack of word inflection, (2) An omission of punctuation and individual words, and (3) Syntactic diversity.

We, therefore, propose the corresponding three heuristics to generate pseudo-glosses from spoken language: (1) Lemmatization of spoken words; (2) POS-dependent and random word deletion; (3) Random word permutation.

We use spaCy \cite{spacy2} for (1) lemmatization and (2) POS tagging to only keep nouns, verbs, adjectives, adverbs, and numerals. We also drop the remaining tokens with probability $p=0.2$, and (3) randomly reorder tokens with maximum distance $d=4$.

\paragraph{Language-specific rules}
While random permutation allows some degree of robustness to word order, it cannot capture all aspects of syntactic divergence between signed and spoken language.
Therefore, inspired by previous work on rule-based syntactic transformations for reordering in MT \citep{collins-etal-2005-clause,isozaki2010head,zhou-etal-2019-handling}, we manually devise a shortlist of syntax transformation rules based on the grammar of DGS and German. %\am{For ASL and English, the ASLG-PC12 dataset \cite{othman2012english} was constructed in a rule-based fashion, and so we use it for the language-specific setting.}

We perform lemmatization and POS filtering as before. In addition, we apply compound splitting \cite{tuggener2016incremental} on nouns and only keep the first noun, reorder German SVO sentences to SOV, move adverbs and location words to the start of the sentence, and move negation words to the end. We provide a detailed list of rules in Appendix \ref{appendix:rules}.

\section{Experimental Setting}\label{sec:experiments}

\subsection{Datasets}\label{sec:datasets}

\paragraph{DGS \& German}
RWTH-PHOENIX-Weather 2014T \cite{Camgoz_2018_CVPR} is a parallel corpus of 8,257 DGS interpreted videos from the \hyperlink{https://www.phoenix.de/}{Phoenix}\footnote{\hyperlink{https://www.phoenix.de/}{www.phoenix.de}} weather news channel, with corresponding SL glosses and German translations.

To obtain monolingual German data, we crawled \hyperlink{https://www.tagesschau.de/}{tagesschau}\footnote{\hyperlink{https://www.tagesschau.de/}{www.tagesschau.de}} and extracted news caption files containing the word ``wetter'' (German for ``weather''). We split the 1,506 caption files into 341,023 German sentences using the spaCy sentence splitter and generate synthetic glosses using our methods described in \S\ref{sec:methods}.

\paragraph{ASL \& English}
The NCSLGR dataset \cite{ncslgr} is a small, general domain dataset containing 889 ASL videos with 1,875 SL glosses and English translations.

We use ASLG-PC12 \cite{othman2012english}, a large synthetic ASL gloss dataset created from English text using rule-based methods with 87,710 publicly available examples, for our experiments on ASL-English with language-specific rules.
We also create another synthetic variation of this dataset using our proposed general rule-based augmentation.

\subsection{Baseline Setup}
We first train a \textbf{Baseline} system on the small manually annotated SL dataset we have available in each language pair.
The model architecture and training method are based on \citet{yin2020better}'s Transformer gloss-to-text translation model.
While previous work (\textbf{\citeauthor{yin2020better} Reimpl.}) used word-level tokenization, for Baseline and all other models described below, we instead use BPE tokenization (\citet{sennrich2015neural}; with 2,000 BPE codes) for efficiency and simple handling of unknown words.
For all tested methods, we repeat every experiment 3 times to account for variance in training.

\subsection{Pre-training on Augmented Data}
For \textbf{General-\emph{pre}} and \textbf{Specific-\emph{pre}}, we pre-train a tokenizer and translation model on pseudo-parallel data obtained using general and language-specific rules respectively, until the accuracy on the synthetic validation set drops. We test both models on the parallel SL dataset in a zero-shot setting.

For \textbf{BT-\emph{tuned}},  \textbf{General-\emph{tuned}} and \textbf{Specific-\emph{tuned}}, we take models pre-trained on pseudo-parallel data obtained with either back-translation, general rules, or language-specific rules, and continue training with half of the training data taken from the synthetic pseudo-parallel data and the other half taken from the real SL data. Then, we fine-tune these models on the real SL data and evaluate them on the test set.

%To better understand the interaction between annotated corpus size and augmentation/pre-training methods, we take a random sub-sample of sentences from each dataset.

\begin{figure}[t]
    \begin{center}
    \begin{subfigure}{0.49\columnwidth}
        \includegraphics[width=1.08\columnwidth]{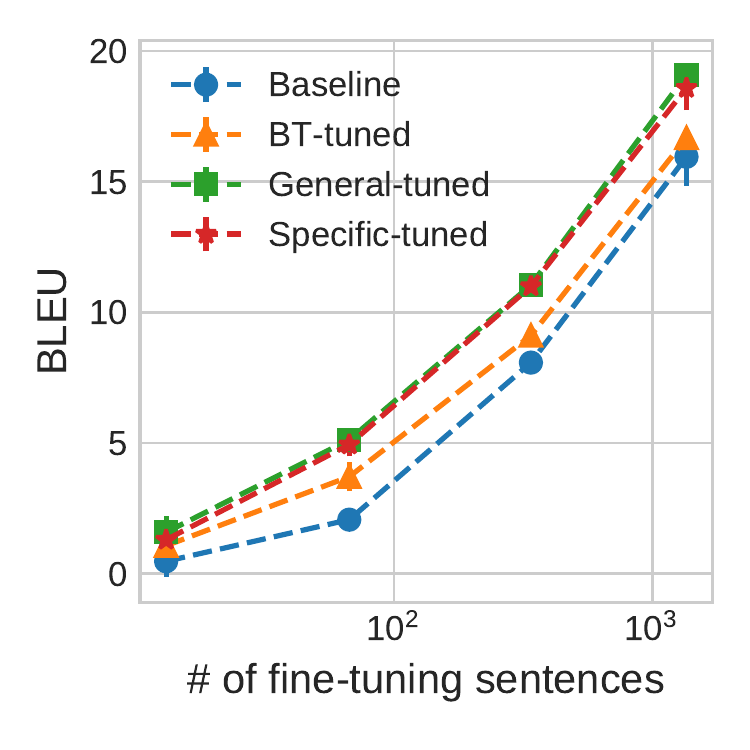}
        \caption{NCSLGR (ASL)}
        \label{fig:res-ncslgr}
    \end{subfigure}
    \begin{subfigure}{0.49\columnwidth}
        \includegraphics[width=1.08\columnwidth]{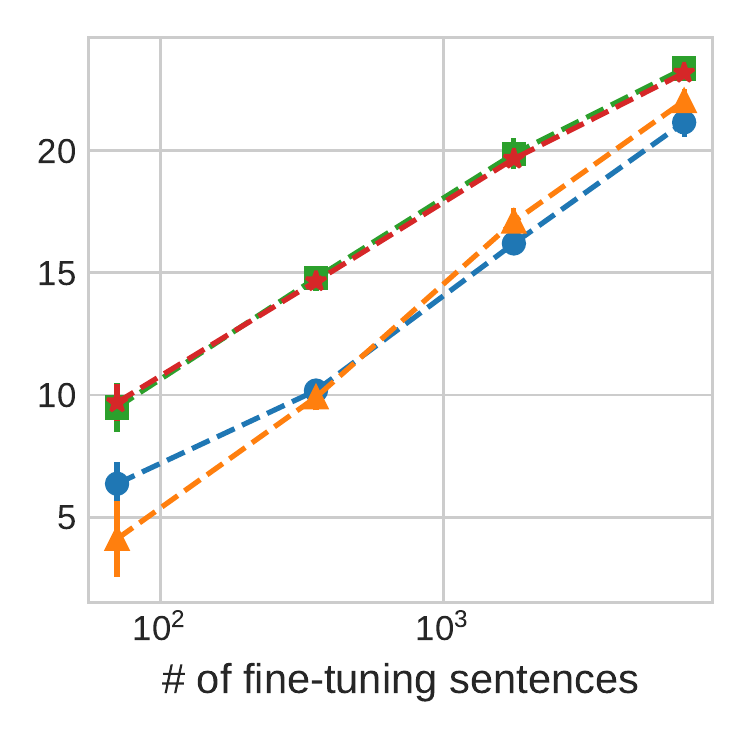}
        \caption{PHOENIX (DGS)}
        \label{fig:res-phoenix}
    \end{subfigure}
    \end{center}
    \caption{Translation results using various amounts of annotated parallel data.}
    \label{fig:res-data}
\end{figure}

\section{Results}

We evaluate our models across all datasets and sizes using SacreBLEU (v1.4.14) \cite{post-2018-call} and COMET (\emph{wmt-large-da-estimator-1719}) \cite{rei-etal-2020-comet}. We also compare our results to previous work on PHOENIX in Table \ref{table:results}. Detailed scores for each experiment are provided in Appendix \ref{appendix:allscores}.

First, we note results on General-\emph{pre} and Specific-\emph{pre}.
Interestingly, the scores are non-negligible, demonstrating that the model can learn with \emph{only} augmented data.%
\footnote{In contrast, merely outputting the source sentence results in 1.36 BLEU, -90.28 COMET on PHOENIX and 1.5 BLEU, -119.45 COMET on NCSLGR.}
Moreover, on PHOENIX Specific-\emph{pre} achieves significantly better performance than General-\emph{pre}, which suggests our hand-crafted syntax transformations effectively expose the model to the divergence between DGS and German during pre-training.

Next, turning to the \emph{tuned} models, we see that Specific and General outperform both the baseline and BT by large margins, demonstrating the effectiveness of our proposed methods.
Interestingly, General-\emph{tuned} performs slightly better, in contrast to the previous result.
We posit that, similarly to previously reported results on sampling-based back translation \citep{edunov-etal-2018-understanding}, General is benefiting from the diversity provided by sampling multiple reordering candidates, even if each candidate is of lower quality.

Looking at Figure \ref{fig:res-data}, we see that the superior performance of our methods holds for all data sizes, but it is particularly pronounced when the parallel-data-only baseline achieves moderate BLEU scores in the range of 5-20.
This confirms that BT is not a viable data augmentation method when parallel data is not plentiful enough to train a robust back-translation system.

% The two models also outperform BT-\emph{tuned}, confirming that by exploiting the lexical overlap between sign and spoken languages, we can mitigate extremely low-resource settings using simple rules on monolingual spoken language text instead of back-translation. 

\begin{table}[t]
\resizebox{\linewidth}{!}{  
\begin{tabular}{lcccc}
\toprule
&\multicolumn{2}{c}{PHOENIX} & \multicolumn{2}{c}{NCSLGR} \\
 & BLEU$\uparrow$ & COMET$\uparrow$  & BLEU$\uparrow$ & COMET$\uparrow$ \\ 
\midrule
% \citet{Camgoz_2018_CVPR} & 19.26 & -- \\ 
% \citet{camgoz2020sign} & 24.54 & -- \\ 
% \citet{yin2020better} & 23.32 & -- \\ \midrule
\citeauthor{yin2020better} Reimpl.\footnotemark  & 22.17 & -2.93 &-&-\\ 
Baseline & 21.15 & -5.74 &15.95&-61.00\\ \midrule
General-\emph{pre} (0-shot) & 3.95 & -69.09 & 0.97 & -135.99 \\
Specific-\emph{pre} (0-shot) & 7.26 & -53.14 & 0.95 & -134.13 \\ \midrule
% BT \emph{pre-tuned} & 22.47 & 3.91 \\ 
BT-\emph{tuned} & \textbf{22.02} & \textbf{6.84} & 16.67 & \textbf{-51.86} \\
% Rules \emph{pre-tuned} & 23.03 & 9.84 \\
General-\emph{tuned} & \textbf{23.35} & \textbf{13.65}& \textbf{19.09} & \textbf{-34.50} \\ 
Specific-\emph{tuned} & \textbf{23.17} & \textbf{11.70} &\textbf{18.5}8&\textbf{-39.96}\\ \bottomrule
\end{tabular}
}
\caption{Results of our different models on PHOENIX and NCSLGR. We \textbf{bold} scores statistically significantly higher than baseline at the 95\% confidence level.}
\label{table:results}
\end{table}
\footnotetext{The original work achieves 23.32 BLEU; correspondence with the authors has led us to believe that the discrepancy is due to different versions of the underlying software.}

\section{Implications and Future Work}

Consistent improvements over the baseline across two language pairs by our proposed rule-based augmentation strategies demonstrate that data augmentation using monolingual spoken language data is a promising approach for sign language translation. 

Given the efficiency of our general rules compared to language-specific rules, future work may also include a more focused approach on specifically pre-training the target-side decoder with spoken language sentences so that by learning the syntax of the target spoken language, it can generate fluent sentences from sign language glosses having little to no parallel examples during training.

% \gn{Overall, the discussion of these results is quite brief, what are the interesting conclusions here? Also, I wonder if there are any interesting features of the results where data augmentation helps (or not helping) in particular types of sentences or for particular types of words, etc. You could try running compare-mt to do a quick analysis and see if anything is interesting there.}
% \am{compare-mt: https://nlp.biu.ac.il/~amit/tmp/compare-mt/ where sys1-7 are our systems from table 2}

% \section{Acknowledgments}
% Do not forget to thank Valentina for helping with the rules

\clearpage

\bibliographystyle{acl_natbib}
\bibliography{anthology,acl2021}

\clearpage

\appendix
\appendixpage
\addappheadtotoc

\begin{table*}[hb]

\resizebox{\linewidth}{!}{
\begin{tabular}{llllllllll}
\toprule
\multicolumn{2}{l}{\% of available annotated data used}  & \multicolumn{2}{l}{1\%} & \multicolumn{2}{l}{5\%} & \multicolumn{2}{l}{25\%} & \multicolumn{2}{l}{100\%} \\
\cmidrule{3-10}
 &  & BLEU & COMET & BLEU & COMET & BLEU & COMET & BLEU & COMET \\
\cmidrule{3-10}
\multirow{4}{*}{PHOENIX} & Baseline & 6.37 $\pm$ 0.89 & -89.21 $\pm$ 12.82 & 10.18 $\pm$ 0.40 & -71.37 $\pm$ 2.86 & 16.20 $\pm$ 0.27 & -33.88 $\pm$ 4.35 & 21.15 $\pm$ 0.58 & -5.74 $\pm$ 2.35 \\
 & BT-\emph{tuned} & 4.12 $\pm$ 1.55 & -91.87 $\pm$ 16.35 & 9.91 $\pm$ 0.54 & \textbf{-53.38 $\pm$ 4.04} & \textbf{17.10 $\pm$ 0.56} & \textbf{-16.46 $\pm$ 2.52} & \textbf{22.02 $\pm$ 0.50} & \textbf{6.84 $\pm$ 0.34} \\
 & General-\emph{tuned} & \textbf{9.49 $\pm$ 1.01} & \textbf{-52.23 $\pm$ 6.31} & \textbf{14.78 $\pm$ 0.51} & \textbf{-27.13 $\pm$ 2.29} & \textbf{19.86 $\pm$ 0.64} & \textbf{-0.72 $\pm$ 2.44} & \textbf{23.35 $\pm$ 0.22} & \textbf{13.65 $\pm$ 1.68} \\
 & Specific-\emph{tuned} & \textbf{9.70 $\pm$ 0.75} & \textbf{-55.94 $\pm$ 2.08} & \textbf{14.65 $\pm$ 0.29} & \textbf{-30.85 $\pm$ 1.45} & \textbf{19.66 $\pm$ 0.08} & \textbf{-5.62 $\pm$ 0.51} & \textbf{23.17 $\pm$ 0.30} & \textbf{11.70 $\pm$ 1.20} \\
\midrule
\multirow{4}{*}{NCSLGR} & Baseline & 0.47 $\pm$ 0.60 & -153.90 $\pm$ 11.89 & 2.07 $\pm$ 0.32 & -145.14 $\pm$ 1.15 & 8.07 $\pm$ 0.43 & -101.24 $\pm$ 5.14 & 15.95 $\pm$ 1.11 & -61.00 $\pm$ 6.86 \\
 & BT-\emph{tuned} & 1.07 $\pm$ 0.47 & \textbf{-139.80 $\pm$ 3.78} & \textbf{3.71 $\pm$ 0.55} & \textbf{-117.33 $\pm$ 3.03} & \textbf{9.11 $\pm$ 0.05} & \textbf{-82.41 $\pm$ 2.29} & 16.67 $\pm$ 0.32 & \textbf{-51.86 $\pm$ 0.66} \\
 & General-\emph{tuned} & 1.58 $\pm$ 0.60 & \textbf{-134.22 $\pm$ 1.73} & \textbf{5.13 $\pm$ 0.15} & \textbf{-106.59 $\pm$ 1.56} & \textbf{11.04 $\pm$ 0.04} & \textbf{-66.35 $\pm$ 2.00} & \textbf{19.09 $\pm$ 0.20} & \textbf{-34.50 $\pm$ 1.19} \\
 & Specific-\emph{tuned} & 1.30 $\pm$ 0.52 & \textbf{-128.14 $\pm$ 1.58} & \textbf{4.94 $\pm$ 0.45} & \textbf{-107.60 $\pm$ 4.01} & \textbf{10.99 $\pm$ 0.12} & \textbf{-71.37 $\pm$ 1.01} & \textbf{18.58 $\pm$ 0.84} & \textbf{-39.96 $\pm$ 1.91} \\ \bottomrule

\end{tabular}
}

\caption{Mean and standard deviation of BLEU and COMET over different experimental settings. We \textbf{bold} scores statistically significantly higher than baseline at the 95\% confidence level.}
\label{table:all-results}
\end{table*}

\section{Data Augmentation Rules}\label{appendix:rules}

\subsection{General Rules}
For a given sentence $\mathcal{S}$:
\begin{enumerate}
    \item Discard all tokens $t \in \mathcal{S}$ if \textbf{POS}$(t) \not\in$ \{noun, verb, adjective, adverb, numeral\}
    \item Discard remaining tokens $t \in \mathcal{S}$ with probability $p = 0.2$
    \item Lemmatize all tokens $t \in \mathcal{S}$
    \item Apply a random permutation $\sigma$ to $\mathcal{S}$ verifying $\forall i \in \{1,n\}, |\sigma(i)-i| \leq 4$
\end{enumerate}
where $n$ is the number of tokens in $\mathcal{S}$ at step 4 and \textbf{POS} is a part-of-speech tagger.

\subsection{German-DGS Rules}
For a given sentence $\mathcal{S}$:
\begin{enumerate}
    \item For each subject-verb-object triplet $(s,v,o) \in \mathcal{S}$, swap the positions of $v$ and $o$ in $\mathcal{S}$
    \item Discard all tokens $t \in \mathcal{S}$ if \textbf{POS}$(t) \not\in$ \{noun, verb, adjective, adverb, numeral\}
    \item For $t \in \mathcal{S}$, if \textbf{POS}$(t) =$ adverb, then move $t$ to the start of $s$
    \item For $t \in \mathcal{S}$, if \textbf{NER}$(t) =$ location, then move $t$ to the start of $s$
    \item For $t \in \mathcal{S}$, if \textbf{DEP}$(t) =$ negation, then move $t$ to the end of $s$
    \item For $t \in \mathcal{S}$, if $t$ is a compound noun $c_1c_2...c_n$, replace $t$ by $c_1$ 
    \item Lemmatize all tokens $t \in \mathcal{S}$
\end{enumerate}
where \textbf{POS} is a part-of-speech tagger, \textbf{NER} is a named entity recognizer and \textbf{DEP} is a dependency parser.

% \begin{algorithm}
% \SetAlgoLined
% \SetKwProg{With}{with}{ do}{end}
% \SetKwInput{Param}{Parameters} 
% \KwIn{\textit{sentence} }
% \Param{\textit{result}(str)}
%  \For{\textit{word} \textup{\textbf{in}} \textit{sentence}}{
%   \If{\textit{word}  \textup{is noun, verb, adjective, adverb or numeral}}{
%   \With{\textup{probability} \textit{p = 0.2}}{\textit{result} += Lemmatizer(\textit{word})}
%   }
%  }
%  apply random permutation $\theta$ such that 
%  \caption{General Rules}
% \end{algorithm}

\newpage

\section{Model Reproduction}\label{appendix:reproduction}

For reproduction purposes, here we lay the exact commands for training a single model using OpenNMT 1.2.0 \cite{klein-etal-2017-opennmt}. These commands are taken from \cite{yin2020better}.

Given 6 files---\emph{train.gloss} / \emph{train.txt}, \emph{dev.gloss} / \emph{dev.txt}, \emph{test.gloss} / \emph{test.txt}---we start by preprocessing the data using the following command:

\begin{lstlisting}[language=bash]
onmt_preprocess \
-dynamic_dict -save_data processed_data \  
-train_src train.gloss -train_tgt train.txt \ 
-valid_src dev.gloss -valid_tgt dev.txt
\end{lstlisting}

Then, we train a translation system using the train command:
\begin{lstlisting}[language=bash]
onmt_train -data processed_data -save_model model \
-layers 2 -rnn_size 512 -word_vec_size 512 -heads 8 \
-encoder_type transformer -decoder_type transformer \
-position_encoding -transformer_ff 2048 -dropout 0.1 \
-early_stopping 3 -early_stopping_criteria accuracy ppl \
-batch_size 2048 -accum_count 3 -batch_type tokens \
-max_generator_batches 2 -normalization tokens \
-optim adam -adam_beta2 0.998 -decay_method noam \ 
-warmup_steps 3000 -learning_rate 0.5 -max_grad_norm 0 \ 
-param_init 0 -param_init_glorot -label_smoothing 0.1 \ 
-valid_steps 100 -save_checkpoint_steps 100 \
-world_size 1 -gpu_ranks 0
\end{lstlisting}

At the end of the training procedure, it prints to console ``Best model found at step X''.
Locate it, and use it for translating the data:

\begin{lstlisting}
onmt_translate -model model_step_X.pt -src test.gloss \
-output hyp.txt -gpu 0 -replace_unk -beam_size 4
\end{lstlisting}

Finally, evaluate the output using SacreBLEU:
\begin{lstlisting}
cat hyp.txt | sacrebleu test.txt
\end{lstlisting}

\section{Full Experimental Results}\label{appendix:allscores}
Table \ref{table:all-results} includes the evaluation scores for all of our experiments, ran three times.

\end{document}